\documentclass[letterpaper]{article} 
\usepackage{aaai2026}  
\usepackage{times}  
\usepackage{helvet}  
\usepackage{courier}  
\usepackage[hyphens]{url}  
\usepackage{graphicx} 
\usepackage{natbib}  
\usepackage{caption} 
\usepackage{algorithm}
\usepackage{algorithmic}
\usepackage{algorithm}
\usepackage{algorithmic}
\usepackage{amsmath}
\usepackage{booktabs}
\usepackage{multirow}
\usepackage{cleveref}
\usepackage{amssymb}

\usepackage{newfloat}
\usepackage{listings}
\DeclareCaptionStyle{ruled}{labelfont=normalfont,labelsep=colon,strut=off} 
\floatstyle{ruled}
\newfloat{listing}{tb}{lst}{}
\floatname{listing}{Listing}
\nocopyright
\title{Towards Non-Stationary Time Series Forecasting with Temporal Stabilization and Frequency Differencing}
\author{
    Junkai Lu\textsuperscript{\rm 1, \rm 2}, Peng Chen\textsuperscript{\rm 1}, Chenjuan Guo\textsuperscript{\rm 1}, Yang Shu\textsuperscript{\rm 1}, Meng Wang\textsuperscript{\rm 2}, Bin Yang\textsuperscript{\rm 1}\thanks{Corresponding Author}
}
\affiliations{
    \textsuperscript{\rm 1}School of Data Science and Engineering, East China Normal University, Shanghai, China\\
    \textsuperscript{\rm 2}School of Computer Science, Xi'an Polytechnic University, Shaanxi, China\\
    \{jklu, pchen\}@stu.ecnu.edu.cn, \{cjguo, yshu, byang\}@dase.ecnu.edu.cn, wangmeng@xpu.edu.cn
}

\usepackage{bibentry}

\begin{document}

\maketitle

\begin{abstract}
Time series forecasting is critical for decision-making across dynamic domains such as energy, finance, transportation, and cloud computing. However, real-world time series often exhibit non-stationarity, including temporal distribution shifts and spectral variability, which pose significant challenges for long-term time series forecasting. In this paper, we propose \textbf{\emph{DTAF}}, a dual-branch framework that addresses non-stationarity in both the temporal and frequency domains. For the temporal domain, the Temporal Stabilizing Fusion (TFS) module employs a non-stationary mix of experts (MOE) filter to disentangle and suppress temporal non-stationary patterns while preserving long-term dependencies. For the frequency domain, the Frequency Wave Modeling (FWM) module applies frequency differencing to dynamically highlight components with significant spectral shifts. By fusing the complementary outputs of TFS and FWM, DTAF generates robust forecasts that adapt to both temporal and frequency domain non-stationarity. Extensive experiments on real-world benchmarks demonstrate that DTAF outperforms state-of-the-art baselines, yielding significant improvements in forecasting accuracy under non-stationary conditions. All codes are available at https://github.com/decisionintelligence/DTAF.
\end{abstract}


\section{Introduction}

With the acceleration of digital transformation, time series have been generated continuously across various domains, including energy \cite{qiu2025dag,wu2025k2vae,li2025TSFM-Bench}, finance \cite{ye2025aligning,liu2025astgi,wu2025enhancing}, transportation \cite{gao2025ssdts,qiu2025multi,wu2024catch}, and cloud computing \cite{hu2024multirc,qiu2025DBLoss,AutoCTS++}. Accurate forecasting of future values from historical time series is essential for decision-making and operational planning~\cite{wang2025lightgts}. 

\begin{figure}[t]   
    \centering
    \includegraphics[width=\linewidth,scale=0.8]{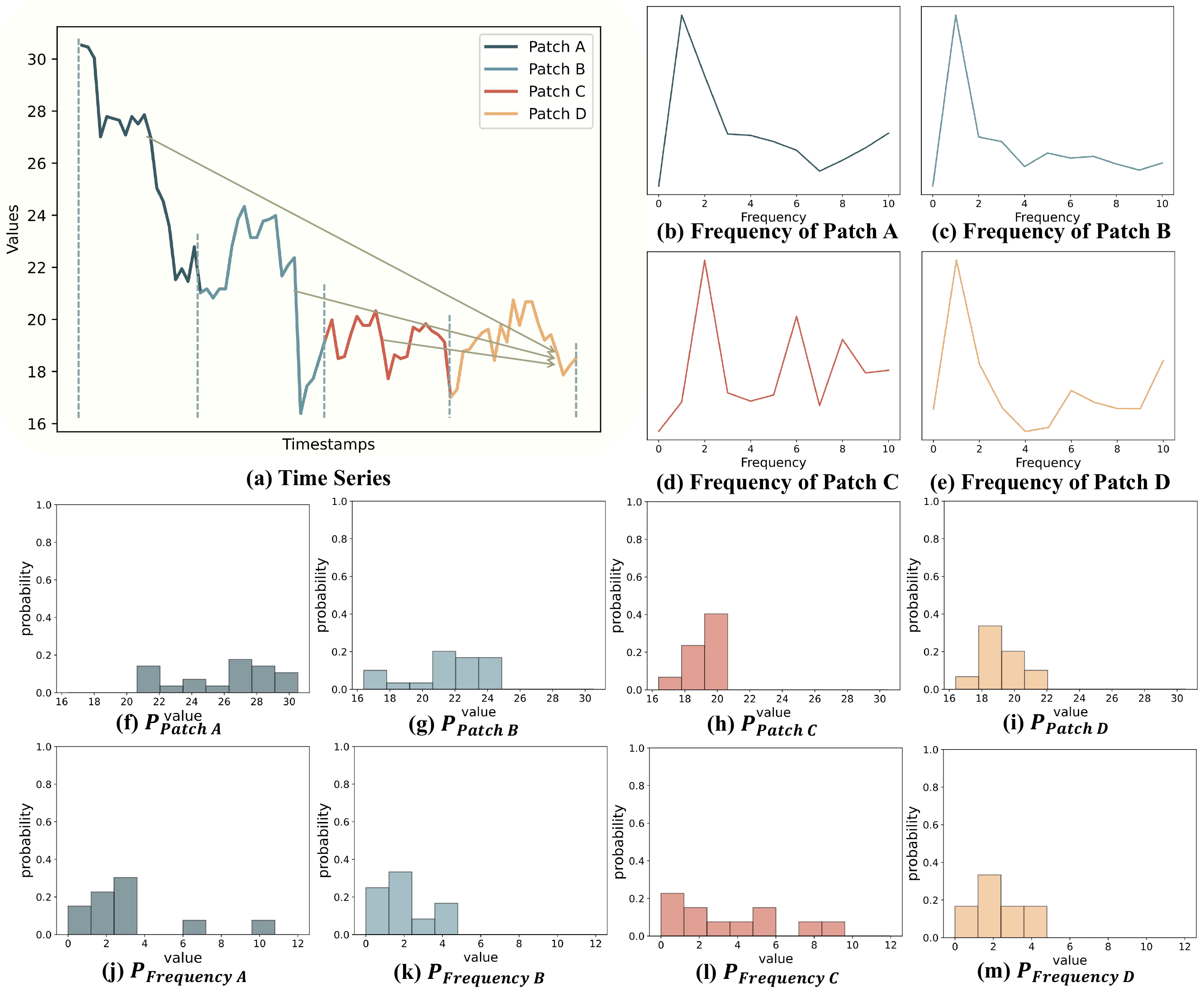}
    \caption{A non-stationary time series split into four patches (A, B, C, D), (b-e) shows the amplitude spectrum of the four patches, (f-i) shows diverse distributions in the temporal domain, and (j-m) shows diverse distributions in the frequency domain, reflecting non-stationarity across both domains.}
    \label{fig:intro}
\end{figure}

In real-world scenarios, time series often exhibit non-stationary patterns in both the temporal and frequency domains~\cite{qiu2025comprehensive, wangtowards}. As illustrated in \Cref{fig:intro}, the entire sequence is divided into four patches, each characterized by distinct patterns. The distributions of these patches evolve over time in temporal and frequency domains, as illustrated in \Cref{fig:intro}(f–i) and \Cref{fig:intro}(j–m), reflecting the underlying non-stationarity~\cite{qiu2025duet}.
These observations highlight the necessity and difficulty of designing methods that can effectively handle non-stationarity in both domains. Specifically, addressing this problem involves overcoming two key challenges:

\textbf{Extracting and disentangling heterogeneous non-stationary patterns to adapt to long-term dependency modeling is challenging}. Time series data inherently consists of stationary and non-stationary components in the temporal domain~\cite{qiu2025easytime,wu2025unlocking, chen2025cc}. Directly modeling long-term temporal dependencies on raw sequences could be severely impaired by non-stationary dynamics, leading to degraded prediction performance. A promising solution involves explicitly modeling the stationary component and isolating non-stationary effects. However, time series non-stationary patterns are highly complex and heterogeneous, making it impossible for a single architecture model to comprehensively model these non-stationary patterns. Therefore, how to design an architecture capable of extracting and disentangling these heterogeneous non-stationary patterns to enable long-term temporal dependency modeling is a challenge.

\textbf{Dynamic modeling frequency drifts and spectral variations present a challenge}. Time series contains rich frequency characteristics that reflect periodicity and trends. These frequency characteristics are inherently non-stationary, changing over time (as shown in \Cref{fig:intro} (j-m)). 
However, existing frequency analysis methods such as the Fourier Transform \cite{fft1} assume stationarity within the observation window and produce global frequency representations. Therefore, they fail to capture transient frequency shifts and specific local patterns. This limitation is particularly evident in applications like electricity load forecasting, where frequency components corresponding to daily and seasonal cycles change over time due to user behavior and environmental factors (e.g., climate changes). In such cases, relying on static frequency analysis may obscure important variations necessary for accurate prediction. Therefore, efficient frequency modeling that can dynamically track frequency drifts is essential.

To address these challenges, we propose a novel Dual-branch Modeling with Temporal Fusion and Frequency Differencing (\textbf{DTAF}), to address the temporal and spectral non-stationarity inherent in time series. As illustrated in \Cref{fig:framework}, DTAF integrates two core components: \textit{Temporal Stabilizing Fusion} (TFS) and \textit{Frequency Wave Modeling} (FWM).

In the temporal domain, we introduce a Non-stationary MOE Filter in TFS, composed of expert networks specialized in distinct non-stationary patterns. These extracted components are subtracted from the input to yield a stationary residual, enabling more robust modeling. Based on it, we conduct temporal fusion to capture long-term dependencies.

In the frequency domain, FWM addresses frequency non-stationarity via spectral differencing, highlighting components with the most significant changes. This allows FWM to adapt to dynamic spectral patterns. Finally, Dual-branch Attention fuses the outputs of TFS and FWM, enabling robust forecasting that accounts for the two domain shifts. Our contributions are summarized as follows:  
\begin{itemize}
    \item To address the challenge of non-stationarity in time series forecasting, we propose a novel model, DTAF, which leverages dual-branch modeling to capture both temporal and frequency-domain non-stationarity.
    \item We propose the Temporal Stabilizing Fusion (TFS) module, integrating a non-stationary MOE Filter, which automatically learns diverse temporal non-stationary patterns, then performs global fusion to capture long-term dependency, thus significantly enhancing prediction.
    \item We introduce the Frequency Wave Modeling (FWM) module, which applies differencing techniques in the frequency domain to enhance the model's ability to model and adapt to frequency domain shifts.
    \item Extensive experiments on eleven real-world datasets have demonstrated that DTAF achieves state-of-the-art prediction accuracy. 
\end{itemize}

\section{Related Work}

\subsection{Multivariate Time Series Forecasting}
Multivariate time series forecasting aims to predict future values of multiple channels based on their historical data \cite{qiu2025comprehensive, cheng2023weakly, wang2023hierarchical}. Traditional methods such as RNN \cite{RNN, ts-RNN} and LSTM \cite{lstm} capture time dependencies, but as the sequence length increases, the historical patterns will be lost. To address these challenges, deep learning approaches such as DeepAR \cite{DeepAR} have emerged, utilizing autoregressive probabilistic models to enhance robustness and accuracy in forecasting. Recent advancements have introduced Graph Neural Networks (GNNs) \cite{enhanceNet, gnn, tpgnn} to model spatial dependencies in time series. For example, in tasks such as energy forecasting \cite{ELC, tian2024Air} , finance prediction~\cite{wang2025plug}, and traffic prediction \cite{Rela-traffic}, GNNs are capable of capturing both temporal information and spatial correlations between variables, thereby improving forecasting accuracy. Simultaneously, Transformer-based architectures use attention mechanisms to model long-range dependencies, such as PatchTST \cite{PatchTST}, Pathformer \cite{pathformer}.


\subsection{Non-Stationary Modeling for Time Series}

Although stationarization is essential for enhancing the predictability of time series data, real-world series invariably exhibit non-stationary patterns. Classical statistical models such as ARIMA \cite{bookTSA} address this by applying temporal differencing to enforce stationarity. In recent years, some methods addressing these challenges on the \textbf{normalization aspect}: Adaptive Norm \cite{AdaptiveNorm}, which applies z-score normalization using global training statistics; DAIN \cite{DAIN}; RevIN \cite{InstanceNorm}, which performs two-stage instance normalization on inputs and outputs; and SAN \cite{SAN}, which dynamically normalizes within shorter windows to estimate local means and variances. And others from the \textbf{model aspect} to capture non-stationary patterns, including Non-stationary Transformer \cite{stationary}, DERITS \cite{DERITS}, AEFIN \cite{AEFIN}, NsDiff \cite{NsDiff}, and TimeBridge \cite{timebridge}. AdaRNN \cite{AdaRNN}, which proposes an adaptive RNN to alleviate the impact of non-stationary factors by characterizing and matching distributions; Triformer \cite{Triformer}, which proposes a light-weight approach to enable variable-specific model parameters, making it possible to capture distinct temporal patterns from different variables. 

Although the existing technologies have alleviated non-stationarity to a certain extent, the importance of jointly modeling it from both the temporal and frequency domains is generally ignored. To address this, we propose a new method, DTAF, that models non-stationarity in the temporal and frequency domains, enhancing prediction performance under non-stationary conditions.

\begin{figure*}[t]

\centerline{\includegraphics[width=\textwidth, scale=0.5]{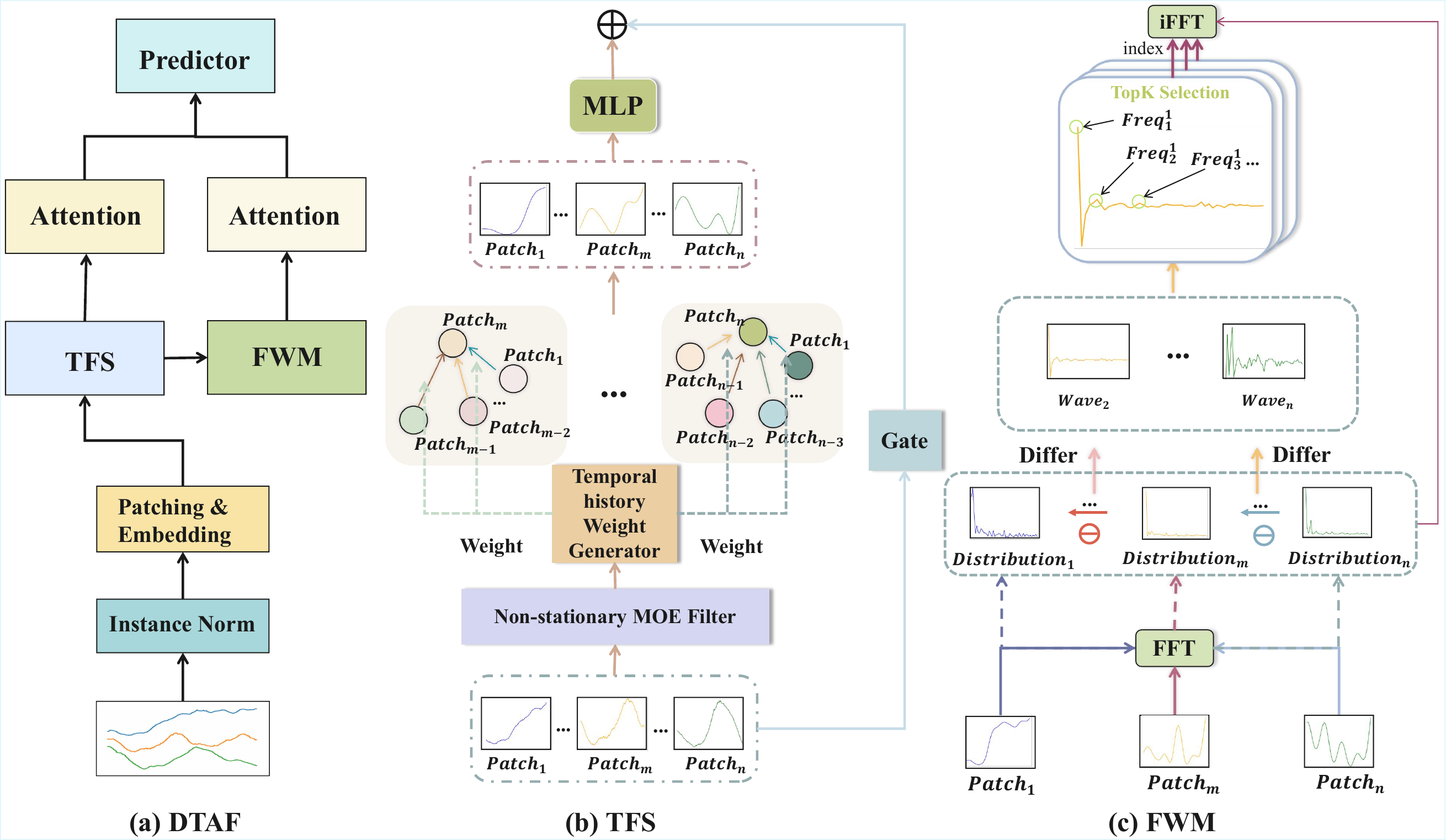}}
\caption{The architecture of DTAF. The two main components of DTAF are TFS and FWM. They are used to aggregate the information in the temporal domain and model in the frequency domain with differencing. Temporal aggregation captures long-term dependencies and local variations, while frequency differencing provides insights into the changing spectral components.}
\label{fig:framework}

\end{figure*}

\section{Methodology}
\subsection{Overall Framework}
To address the non-stationarity in the temporal and frequency domains, we propose a novel model, Dual-branch Modeling with Temporal Fusion and Frequency Differencing (DTAF). As illustrated in \Cref{fig:framework}, DTAF comprises several core components: Instance Norm \cite{InstanceNorm}, Patching\&Embedding \cite{PatchTST}, Temporal Stabilizing Fusion (TFS), Frequency Wave Modeling (FWM), and a Predictor \cite{FC}. Instance Norm alleviates distributional shifts between training and inference, enhancing the robustness of the model against variations in data distributions and improving generalization performance. Inspired by the patching strategy in Transformer architectures \cite{PatchTST}, Patching\&Embedding segments long sequences into patches and embeds them into high-dimensional representations, preserving local temporal patterns. The TFS and FWM modules are specifically designed to mitigate non-stationarity in the temporal and frequency domains. Finally, the Predictor integrates the features from both branches to produce accurate prediction results.

The core of design is TFS and FWM, which are designed to model non-stationarity in the temporal and frequency domains. TFS utilizes a Non-stationary MoE Filter to dynamically extract and remove non-stationary patterns and then uses the Temporal Fusion module to model long-term dependency based on the stationary residual. Complementing the temporal modeling, FWM addresses frequency non-stationarity by applying differencing in the frequency domain, highlighting components with the most significant spectral changes. These enable DTAF to adapt to temporal and frequency dynamics simultaneously, leading to more accurate forecasting under non-stationary conditions. To maintain clarity and focus in the main text, we include essential definitions and task formulations in \Cref{Appendix: dataset}.

\subsection{Temporal Stabilizing Fusion}
\label{model:TFS}

In complex time series, the heterogeneous non-stationary patterns lead to inconsistent distributions among patches, reducing the prediction performance.

As illustrated in \Cref{fig:intro} (f–i), although these patches may follow similar global trends, their local fluctuations vary drastically. These heterogeneous non-stationary patterns make it difficult to effectively capture long-term temporal dependencies.
 
To address this, we introduce the Temporal Stabilizing Fusion module (TFS). It first applies a non-stationary MOE Filter to extract and subtract these non-stationary components, and then performs a temporal fusion module to capture long-term dependencies. By extracting and filtering heterogeneous non-stationary patterns, TFS enables robust long-term modeling and significantly enhances forecasting accuracy under dynamical non-stationary conditions.

\textbf{Non-stationary MOE Filter.} To alleviate the impact of non-stationarity in the temporal modeling, we propose the Non-stationary MOE Filter, which is designed to learn non-stationary patterns and remove them from the original sequence, stabilizing the input and making it more suitable for long-term modeling. Specifically, the Non-stationary MOE Filter consists of a collection of Experts dedicated to extracting non-stationary patterns, along with a routing network that dynamically assigns inputs to the experts. 

For notational clarity, we describe our method using a univariate time series $\mathbf{X} \in \mathbb{R}^{T}$, where $T$ denotes the input sequence length (extension to multivariate cases is straightforward). We first apply instance normalization to standardize the input statistics, mitigating distribution shift. Recognizing that individual time steps often lack meaningful context, we employ a patching strategy that segments the sequence into $N$ overlapping patches $\{\mathbf{X}^1, \mathbf{X}^2, \cdots, \mathbf{X}^N\}$, where each patch captures local temporal patterns. These patches are then projected into a high-dimensional latent space through patch embedding, yielding the embeddings $\{\mathbf{X}_{\text{patch}}^1, \mathbf{X}_{\text{patch}}^2, \cdots, \mathbf{X}_{\text{patch}}^N\}$ for subsequent processing, where $\mathbf{X}_{\text{patch}}^i \in \mathbb{R}^{d}$ represents $i$-th patch embedding, and $d$ denotes the embedding dimension.

For $i$-th input patch embedding \( \mathbf{X}_{\mathrm{patch}}^i \), the routing network first assigns a weight to each expert's contribution based on each input characteristic, and then each expert utilizes special MLP networks to extract potential non-stationary patterns. After aggregating the weighted outputs from all experts, the non-stationary pattern \( \mathbf{X}_{\mathrm{patterns}}^i \in \mathbb{R}^{d} \) for the sequence is obtained, which is subtracted from the input patch \( \mathbf{X}_{\mathrm{patch}}^i \), as illustrated by the following equations:
\begin{equation}
\begin{aligned}
    \mathcal{E} &= \left \{ e_1, e_2, e_3, ..., e_m \right \}, \\
    \mathbf{router}^i &= \mathrm{Softmax}(\mathrm{Linear}(\mathbf{X}_{\mathrm{patch}}^i)), \\
    \mathbf{X}_{\mathrm{patterns}}^i &= \sum_{j=1}^{m} e_j(\mathbf{X}^i) * \mathbf{router}^i_j,
\end{aligned}
\end{equation}
where \( \mathcal{E} \) denotes the expert set and \( \mathrm{Linear}(\cdot) \) denotes a special linear layer \cite{kan}, and each expert denotes multiple linear layers. To ensure that the experts learn the non-stationary patterns, we introduce a KL divergence \cite{KL} loss to enforce this constraint. Specifically, the loss function is formulated as:
\begin{align}
    \mathbf{X}_{\mathrm{stable}}^i &= \mathbf{X}_{\mathrm{patch}}^i - \mathbf{X}_{\mathrm{patterns}}^i, \\
    \mathcal{L}_{\mathrm{stable}} &= \alpha * \sum_{i=1}^{N} \sum_{j=1}^{N} \mathrm{KL}(\mathbf{X}_{\mathrm{stable}}^i, \mathbf{X}_{\mathrm{stable}}^j) / N^2,
\end{align}
where \( \mathrm{KL}(\cdot) \) calculates the KL divergence and \( \alpha \) is a hyperparameter. KL divergence can make the distribution of each output patch stationary. Since the outputs are the features extracted by subtracting exports from the original patches, exports can learn non-stationary features.

\textbf{Temporal Fusion.} After extracting and removing non‑stationary components from each input patch embedding, we obtain stationary patterns to model long-term temporal dependencies. To further enhance robustness against potential residual non-stationary effects, we propose a temporal fusion module that dynamically balances historical context and current information. 

For each patch \(\mathbf{X}_{\mathrm{stable}}^i \in \mathbb{R}^{d}\), a linear feature extractor \( (\mathrm{LinearExtra}(\cdot)) \) that incorporates time series decomposition is first employed to capture the patterns of the current patch \(\mathbf{X}_{\mathrm{stable}}^i\), formulated as follows:  
\begin{align}
\mathbf{X}^i_t = \mathrm{Avgpool}&(\mathrm{padding}(\mathbf{X}_{\mathrm{stable}}^i))),
\mathbf{X}^i_s = \mathbf{X}_{\mathrm{stable}}^i - \mathbf{X}^t_i,\\
\mathbf{X}^i_h &= \mathbf{W}^t_i \cdot \mathbf{X}^t_i + \mathbf{W}^s_i \cdot \mathbf{X}^s_i,
\end{align}
where \(\mathbf{X}^i_t, \mathbf{X}^i_s \in \mathbb{R}^{d}\) represent the decomposed trend and seasonal components respectively \cite{Autoformer}, \(\mathbf{W}^i_t, \mathbf{W}^i_s \in \mathbb{R}^{d \times d}\) are learnable parameters and \(\mathbf{X}^i_h \in \mathbb{R}^{d}\) is the temporal representation of the \(i\)-th patch. Next, a linear layer gets the weights between the current patch and each historical patch, reflecting the degree of information fusion required from each historical patch:  
\begin{align}
\mathbf{Weight}^i &= \operatorname{\mathrm{Softmax}}(\mathbf{W}_{history} \cdot \mathbf{X}^i_h + \mathbf{b}_{history}),
\end{align}
where \(\mathbf{W}_{\mathrm{history}} \in \mathbb{R}^{N \times d}\), and \(\mathbf{b}_{\mathrm{history}} \in \mathbb{R}^N\). To ensure no future information leaks into the model, all weights \(\mathbf{Weight}^i_n\), for \(n > i\), are set to zero, enforcing the use of historical patches exclusively. Based on the weights generated by the Temporal history Weight Generator, we then integrate the historical patch information through a weighted aggregation, formulated as:  
\begin{align}
\mathbf{X}^{i}_{\mathrm{history}}&=\mathrm{MLP}(\sum_{n=1}^{i-1}\mathbf{Weight}^i_n\cdot \mathbf{X}_{\mathrm{stable}}^n),
\end{align}
Where \( \mathrm{MLP}(\cdot) \) is implemented as one linear transformation layer, facilitating the extraction and reinforcement of key history features. To incorporate the current patch’s feature representations, we introduce a gating mechanism designed to dynamically amplify or suppress the contribution of the current patch to the final representations:  
\begin{align}
\mathbf{X}^{i}_{current} &=\mathrm{Gate}(\mathbf{X}_{\mathrm{patch}}^i)\cdot \mathbf{X}_{\mathrm{patch}}^i, \\
\mathrm{Gate}(\mathbf{X}_{\mathrm{patch}}^i) &= \mathbf{W}_{\mathrm{gate}} \cdot \mathrm{LinearExtra}(\mathbf{X}_{\mathrm{patch}}^i) + \mathbf{b}_{gate},
\end{align}
where \( \mathbf{W}_{gate} \in \mathbb{R}^{1 \times d} \), \( \mathbf{b}_{\mathrm{gate}} \in \mathbb{R}^{1} \) are learnable parameters. Finally, the fused representation is obtained by combining the current patch’s features with the historical information: 
\begin{align}
\mathbf{H}^{i}_{\mathrm{t}}&=\mathbf{X}^{i}_{\mathrm{current}} + \mathbf{X}^{i}_{\mathrm{history}},
\end{align}
where \( \mathbf{H}^{i}_{\mathrm{t}} \in \mathbb{R}^{d} \) represents global representation to the \(i\)-th patch. By integrating historical and current information, DTAF captures essential global patterns while further suppressing residual non-stationary factors, enabling more effective modeling of long-term temporal dependencies.

Overall, DTAF effectively enhances long-term temporal modeling by jointly leveraging the Non-stationary MOE Filter and the Temporal Fusion module. The MOE Filter extracts heterogeneous non-stationary patterns from each patch, producing relatively stationary representations. 

Based on the above components, the Temporal Fusion module adaptively integrates historical and current information to further adapt to the non-stationarity. This two-stage process stabilizes temporal dynamics and enhances long-term dependency modeling under non-stationary conditions.

\subsection{Frequency Wave Modeling}
\label{model:FWM}
Time series data often reveal trends and seasonal patterns through their frequency characteristics. However, these frequency components are not static, leading to non-stationarity in the frequency domain as shown in \Cref{fig:intro} (j-m). Existing approaches typically assume that the frequency patterns of time series remain stable over time. However, such assumptions fail to account for the non-stationarity in the frequency domain, where key components may shift over time. This limits the model’s ability to capture evolving seasonal or trend patterns in time series forecasting.

To address this limitation, we propose a novel module named Frequency Wave Modeling, which is specifically designed to model non-stationarity in the frequency domain. By dynamically tracking the temporal changes of frequency components and selectively emphasizing those that exhibit significant variations, the module enables the model to concentrate on the most informative spectral patterns. 

Based on the patch representation $\mathbf{H}_{t}^i$ from the TFS module, we introduce Frequency Wave Modeling to further characterize non-stationary patterns in the frequency domain as shown in \Cref{fig:framework} (c). First, similar to other approaches \cite{timesnet}, we apply the Fast Fourier Transform (FFT) to each patch, as expressed as:
\begin{equation}
\mathbf{Freq}^i = \mathrm{rFFT}(\mathbf{H}_{\mathrm{t}}^i),
\end{equation}
where \(\mathrm{rFFT}(\cdot)\) denotes the real-valued Fast Fourier Transform that converts time-domain information into the frequency domain, yielding a set of spectral components. These spectral components capture the distribution of frequency components at different patches. Subsequently, we perform a differencing operation on the spectrum sequence and select the frequency components exhibiting significant variation, which can emphasize non-stationary components:
\begin{align}
\mathbf{Wave}^i &= \mathbf{Freq}^i - \mathbf{Freq}^{i-1}(i\geq 2), \\
\mathbf{Picks}^{i} &= \mathrm{Topk}(\mathbf{Wave}^{i}), \\
\mathbf{Freq}^i & _{f_j \notin \mathrm{Picks}^{i}} = 0,
\end{align}
where $\mathbf{Wave} \in \mathbb{R}^{N\times d/2+1}$ represents the degree of change of each frequency in each patch, and for the first patch, we set it to $\mathbf{Freq}^1$; The \(\mathrm{Topk}(\cdot)\) function selects the \(k\) most significantly changing frequency components for every patch, thus obtaining $\mathbf{Picks} \in \mathbb{R}^{N\times k}$, which includes the k most changing frequency components for each patch. For any frequency \(f_j\) , which is not be selected, the corresponding \(\mathbf{Freq}^i_{f_j}\) is set to zero. Finally, we apply the inverse Fast Fourier Transform in \( \mathbf{Freq}^i \in \mathbb{R}^{d/2+1} \) to convert the selected frequency features back to the time domain, obtaining the representation \( \mathbf{H}_\mathrm{f} \in \mathbb{R}^{N \times d} \).

By incorporating differencing operators to track evolving spectral patterns over time, FWM constructs discriminative spectral representations that capture complex, time-varying frequency dynamics, enhancing forecasting performance.

\subsection{Dual-branch Attention}

Based on temporal and frequency features, we then fuse the two domain features. This fusion allows the model to leverage both temporal and frequency information, leading to more robust and accurate predictions. Traditional fusion approaches, such as direct concatenation or using cross-attention to reconstruct one domain from the other, often fall short in capturing the complementary nuances between temporal and frequency features. We propose a dual-branch attention feature fusion mechanism, referred to as Dual-branch Attention, which can be formulated as follows:
\begin{align}
\mathbf{Atten_\mathrm{t}} &= \mathrm{Softmax} (\frac{\mathbf{Q}_\mathrm{t} \mathbf{K}_\mathrm{t}^T}{\sqrt{d}})\mathbf{V}_\mathrm{t}, \\ \mathbf{Atten_\mathrm{f}} &= \mathrm{Softmax}(\frac{\mathbf{Q}_\mathrm{f} \mathbf{K}_\mathrm{f}^T}{\sqrt{d}})\mathbf{V}_\mathrm{f}, \\
\mathbf{H}_{\mathrm{fusion}} &= \mathrm{Concat}(\mathbf{Atten_\mathrm{t}}, \mathbf{Atten_\mathrm{f}}).
\end{align}
Here \( \mathbf{Q}_\mathrm{t}, \mathbf{K}_\mathrm{t}, \mathbf{V}_\mathrm{t} \in \mathbb{R}^{N\times d}\) are derived from the temporal-domain features \( \mathbf{H}_\mathrm{t} \in \mathbb{R}^{N \times d} \) through linear transformations, and \( \mathbf{Q}_\mathrm{f}, \mathbf{K}_\mathrm{f}, \mathbf{V}_\mathrm{f} \in \mathbb{R}^{N\times d}\) are derived from the frequency-domain features \( \mathbf{H}_\mathrm{f} \) through similar transformations. After obtaining the attention of the temporal domain \( (\mathbf{Atten_\mathrm{t}} \in \mathbb{R}^{N \times d}) \)
\begin{table*}[htbp]
\setlength{\tabcolsep}{1mm}
\small
\begin{tabular}{cc|cc|cc|cc|cc|cc|cc|cc|cc|cc}
\toprule
\multicolumn{2}{c|}{\multirow{2}{*}{Models}} & \multicolumn{2}{c}{DTAF} & \multicolumn{2}{c}{Amplifier} & \multicolumn{2}{c}{iTransformer} & \multicolumn{2}{c}{Pathformer} & \multicolumn{2}{c}{FITS} & \multicolumn{2}{c}{TimeMixer} & \multicolumn{2}{c}{PatchTST} & \multicolumn{2}{c}{DLinear} & \multicolumn{2}{c}{Stationary} \\
\multicolumn{2}{c|}{} & \multicolumn{2}{c}{(ours)} & \multicolumn{2}{c}{(2025)} & \multicolumn{2}{c}{(2024)} & \multicolumn{2}{c}{(2024)} & \multicolumn{2}{c}{(2024)} & \multicolumn{2}{c}{(2024)} & \multicolumn{2}{c}{(2023)} & \multicolumn{2}{c}{(2023)} & \multicolumn{2}{c}{(2022)} \\
\addlinespace\cline{1-20} \addlinespace
\multicolumn{2}{c|}{Metrics} & mse & mae & mse & mae & mse & mae & mse & mae & mse & mae & mse & mae & mse & mae & mse & mae & mse & mae \\
\midrule

\multicolumn{2}{c|}{ILI}  & \textbf{1.688} & \textbf{0.801} & 1.819 & 0.888 & 1.857 & 0.892 & 1.995& 0.909 & 2.334 & 1.052 & \underline{1.819} & \underline{0.886} & 1.902 & 0.879 & 2.185 & 1.040 & 2.389 & 1.027 \\ 
\addlinespace\cline{1-20} \addlinespace

\multicolumn{2}{c|}{Covid-19} & \textbf{1.351} & \textbf{0.040} & 5.578 & 0.112 & \underline{1.488} & \underline{0.049} & 3.326 & 0.086 & 5.669 & 0.123 & 12.941 & 0.133 & 1.697 & 0.056 & 8.074 & 0.239 & 2.658 & 0.078 \\ 
\addlinespace\cline{1-20} \addlinespace

\multicolumn{2}{c|}{NN5} & \textbf{0.643} & \textbf{0.538} & 1.794 & 1.018 & 0.660 & \underline{0.550} & 0.698 & 0.582 & 0.811 & 0.653 & \underline{0.655} & 0.593 & 0.688 & 0.589 & 0.691 & 0.580 & 1.295 & 0.915\\ 
\addlinespace\cline{1-20} \addlinespace

\multicolumn{2}{c|}{ETTh} & \textbf{0.369} & \textbf{0.398} & 0.385 & 0.416 & 0.404 & 0.425 & 0.388 & 0.410 & \underline{0.371} & \underline{0.406} & 0.388 & 0.419 & 0.384 & 0.415 & 0.447 & 0.453 & 0.521 & 0.493 \\ 
\addlinespace\cline{1-20} \addlinespace

\multicolumn{2}{c|}{ETTm} & \textbf{0.297} & \textbf{0.338} & 0.327 & 0.364 & 0.314 & 0.358 & \underline{0.305} & \underline{0.341} & \underline{0.305} & 0.345 & 0.306 & 0.349 & 0.302 & 0.347 & 0.307 & 0.351 & 0.456 & 0.430 \\ 
\addlinespace\cline{1-20} \addlinespace

\multicolumn{2}{c|}{Weather} & \textbf{0.222} & \textbf{0.250} & \underline{0.222} & 0.263 & 0.232 & 0.269 & 0.225 & \underline{0.257} & 0.243 & 0.280 & 0.226 & 0.264 & 0.223 & 0.261 & 0.242 & 0.293 & 0.293 & 0.315 \\  
\addlinespace\cline{1-20} \addlinespace

\multicolumn{2}{c|}{Electricity} & \textbf{0.160} & \textbf{0.248} & 0.174 & 0.267 & \underline{0.163} & \underline{0.258} & 0.168 & 0.261 & 0.169 & 0.265 & 0.184 & 0.284 & 0.171 & 0.270 & 0.167 & 0.264 & 0.194 & 0.295 \\ 
\addlinespace\cline{1-20} \addlinespace

\multicolumn{2}{c|}{Traffic} & 0.402 & \textbf{0.249} & 0.423 & 0.294 & \textbf{0.397} & 0.281 & 0.416 & \underline{0.264} & 0.429 & 0.302 & 0.409 & 0.279 & \underline{0.397} & 0.275 & 0.418 & 0.287 & 0.621 & 0.339 \\  
\addlinespace\cline{1-20} 
\addlinespace

\multicolumn{2}{c|}{$1^{st}$ Count}& \textbf{7} & \textbf{8} & 0 & 0 & \underline{1} & \underline{0} & 0 & 0 & 0 & 0 & 0 & 0 & 0 & 0 & 0 & 0 & 0 & 0 \\ 
\addlinespace

\bottomrule
\end{tabular}
\caption{Multivariate forecasting results. Results are averaged from all forecasting horizons. The input lengths of ILI, Covid-19, and NN5 are selected from {36, 104}, while the remaining datasets are selected from {96, 336, 512}. The best results are highlighted in bold, and the second-best results are underlined. ETTh and ETTm represent the average of ETTh1 and ETTh2 and the average of ETTm1 and ETTm2. Due to space limitations, only a part is displayed; the full results are listed in Appendix B.7.}
\label{forecasting results avg}
\end{table*}
and the attention of the frequency domain \( (\mathbf{Atten_\mathrm{f}} \in \mathbb{R}^{N \times d}) \), we concatenate them to form \( \mathbf{H}_{\mathrm{fusion}} \in \mathbb{R}^{2N \times d} \). 
The Dual-branch Attention simultaneously extracts features from both the time and frequency domains, providing richer features for subsequent prediction. Finally, the fused features pass through a prediction module (Predictor), which consists of a fully connected layer:

\begin{align}
\mathbf{X}_{\mathrm{out}} &= \mathrm{Linear}(\mathrm{Flatten}(\mathbf{H}_{\mathrm{fusion}})),
\end{align}
where \( \mathbf{X}_{\mathrm{out}} \in \mathbb{R}^{F} \), \( F \) is the prediction length of time series. This module adaptively selects the required features from the two domains to produce the final prediction \( \mathbf{X}_{\mathrm{out}} \).

\section{Experiment}

\subsection{Dataset} We select 11 widely used time-series datasets for our model, covering a range of domains such as electricity (ETTh1, ETTh2, ETTm1, ETTm2, Electricity), health (ILI, Covid-19), traffic (Traffic), web (Wike2000), environment (Weather), and banking (NN5).
The details of the datasets are shown in Appendix B.1.

\subsection{Baselines} We select several state-of-the-art models as baselines and evaluate them in TFB \cite{tfb}, including Amplifier \cite{Amplifier}; Non-stationary Transformer (Stationary) \cite{stationary}; iTransformer \cite{itransformer}; Pathformer \cite{pathformer}; PatchTST \cite{PatchTST}; Crossformer \cite{crossformer}; FITS \cite{FITS}; TimeMixer \cite{timemixer}; TimesNet \cite{timesnet}, and DLinear \cite{FC} and the details of baselines are shown in Appendix B.2.

\subsection{Settings}
For the forecasting horizon, the ILI, Covid-19, NN5, and Wike2000 datasets are evaluated using prediction lengths of {24, 36, 48, 60}, while the remaining datasets employ prediction lengths of {96, 192, 336, 720}. DTAF is optimized using the AdamW optimizer \cite{adamw}. We select MAE and MSE as metrics. The loss is combined with three parts: L1 loss as task loss, r-drop loss \cite{r-drop} as robust loss to improve robustness of the model, and stable loss, which uses KL to constrain the Non-stationary MOE Filter to learn Non-stationary patterns. It can be formulated as: \(\mathcal{L} = \mathcal{L}_{\mathrm{task}} + \alpha \cdot \mathcal{L}_{\mathrm{stable}} + \beta \cdot \mathcal{L}_{\mathrm{robust}}\), where \( \alpha, \beta \) is a hyperparameter. And the evaluation is run in the TFB \cite{tfb} benchmark. All results of the experiments are the average of running three times.

\begin{figure*}[htbp!]   
    \centering
    \includegraphics[width=\linewidth,scale=1.0]{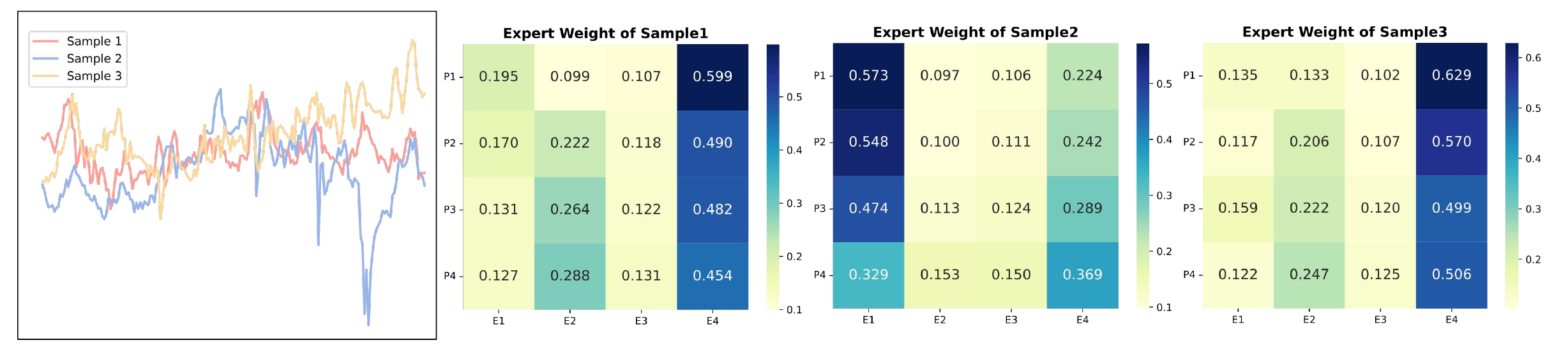}
    \caption{The weights of different experts for the ETTh1. P1, P2, P3, and P4 denote four patches from the sequence, while E1, E2, E3, and E4 represent four experts to extract non-stationary patterns.}
    \label{fig:visual}
\end{figure*}

\subsection{Main Results}

\Cref{forecasting results avg} presents the performance of our model, DTAF, which consistently surpasses existing baselines across two evaluation metrics. 1) \textbf{\textit{Significant Improvements over Baselines}}: Among the \textbf{16} evaluation metrics, DTAF ranks first in \textbf{15} cases. Compared with Stationary, which aims to handle non-stationarity through input transformation, DTAF demonstrates outstanding performance on non-stationary datasets such as Covid-19 (reduces MSE by \textbf{49.0\%} and MAE by \textbf{50.0\%}) and NN5 (reduces MSE by \textbf{50.4\%} and MAE by \textbf{41.2\%}), proving the effectiveness of DTAF in modeling non-stationary data.
2) \textbf{\textit{Robust Predictive Capabilities}}: Compared to other models such as Crossformer, PatchTST, TimeMixer, and other state-of-the-art approaches, DTAF not only achieves notable accuracy improvements but also maintains stable performance across diverse time series datasets, demonstrating strong adaptability to varying data distributions and temporal patterns.

\subsection{Ablation Studies}

In this ablation study, we evaluate DTAF on one non-stationary time series dataset and one normal dataset by disabling each of its three core components: Temporal Stabilizing Fusion (TFS), Frequency Wave Modeling (FWM), and the dual‑branch attention mechanism. As shown in \Cref{tab: Ablation study average results}, the removal of any single module leads to a clear degradation in both MSE and MAE. Without TFS, the model cannot better adapt to the non-stationary patterns while modeling long-term dependencies; omitting FWM undermines its ability to track changing periodic patterns. These findings confirm that each module contributes indispensably to handling different facets of non‑stationarity and that their collaboration is critical to DTAF’s superior forecasting accuracy.

\begin{table}[htbp!]
\centering
\scalebox{0.7}{
\begin{tabular}{cc|cccccccccc}
\toprule
  \multicolumn{2}{c|}{\textbf{Model}} & \multicolumn{2}{c}{\textbf{DTAF}} & \multicolumn{2}{c}{\textbf{w/o  TFS}} & \multicolumn{2}{c}{\textbf{w/o  FWM}} & \multicolumn{2}{c}{\textbf{w/ Attention}} \\
\multicolumn{2}{c|}{Metrics}  & mse & mae & mse & mae & mse & mae & mse & mae\\

\midrule
\multirow[c]{4}{*}{\rotatebox{90}{NN5}} & 24 & \textbf{0.716} & \textbf{0.559} & 0.723 & 0.567 & 0.729 & 0.574 & 0.725 & 0.571  \\ 
 & 36 & \textbf{0.644} & \textbf{0.537}  & 0.659 & 0.550 & 0.659 & 0.554 & 0.660 & 0.554  \\ 
 & 48 & \textbf{0.621} & \textbf{0.535} & 0.623 & 0.540 & 0.632 & 0.549 & 0.635 & 0.552  \\ 
 & 60 & \textbf{0.593} & \textbf{0.522} & 0.608 & 0.535 & 0.597 & 0.527 & 0.608 & 0.536  \\ 
\addlinespace\cline{1-10} \addlinespace

\multirow[c]{4}{*}{\rotatebox{90}{ETTh1}} & 96 & \textbf{0.359} & \textbf{0.384} & 0.367 & 0.391 & 0.363 & 0.391 & 0.376 & 0.395 \\ 
 & 192 & \textbf{0.400} & \textbf{0.414} & 0.408 & 0.421 & 0.407 & 0.419 & 0.418 & 0.430  \\ 
 & 336 & \textbf{0.415} & \textbf{0.424} & 0.426 & 0.433 & 0.420 & 0.434 & 0.438 & 0.445  \\ 
 & 720 & \textbf{0.421} & \textbf{0.449} & 0.457 & 0.468 & 0.446 & 0.460 & 0.450 & 0.463 \\ 

\bottomrule
\end{tabular}
}
\caption{Ablation Study. W/O TFS, W/O FWM, represent removing the TFS and FWM, and W/ Attention represents replacing Dual-branch Attention with Cross Attention. Results averaged over 3 seeds; std $\leq$ ± 0.0058.}

\label{tab: Ablation study average results}
\end{table}
\subsection{Model Analysis}
\label{model_analysis}

\textbf{Visualization of Expert Weights.} We select three samples and show the expert weight distribution for each patch in \Cref{fig:visual}. Our observations reveal that the samples possess unique expert weight distributions. Both samples 1 and 3 share similar seasonality and similar trend patterns, which indicates that they may have similar non-stationarity, thereby yielding similar weight distributions. In contrast, the patterns of sample 2 diverge from samples 1 and 3. These findings underscore the adaptability of the Non-stationary MOE Filter and its ability to extract and distinguish different non-stationary patterns.

\begin{figure}[htbp!]   
    \centering
    \includegraphics[width=\linewidth,scale=0.6]{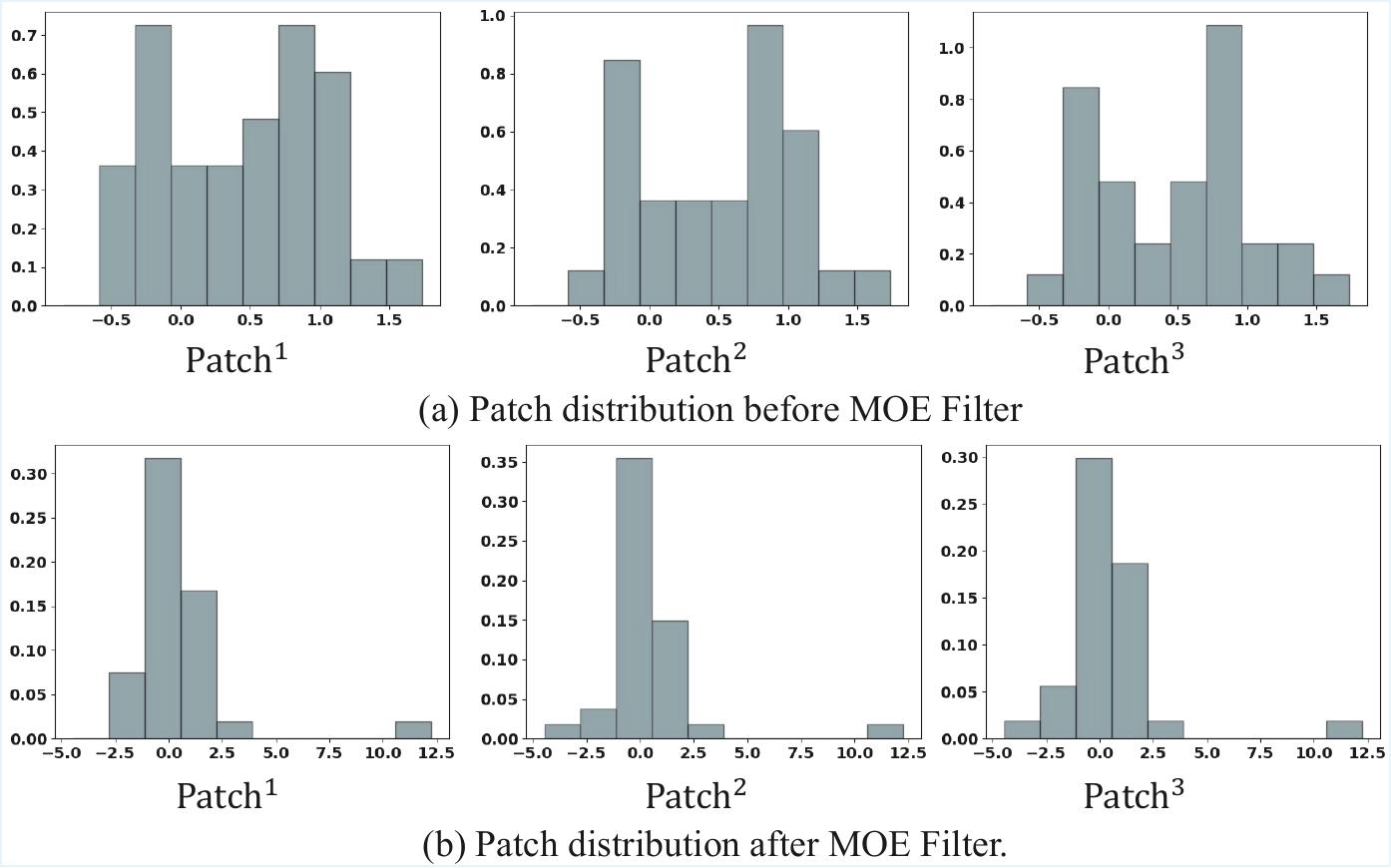}
    \caption{Non-stationary MOE Filter Analysis.}
    \label{fig:moe_case}
\end{figure}

\textbf{Non-stationary MOE Filter Analysis}. We analyze the Non-stationary MOE Filter and carry out a visual analysis of the input and output of the Non-stationary MOE Filter, which is shown in \Cref{fig:moe_case}. It can be seen that before the Non-stationary MOE Filter, the distribution among patches had significant differences. However, after the Filter, the distribution similarity among patches is relatively high. This proves that the Non-stationary MOE Filter can indeed remove the Non-stationary part between patches, making the distribution between patches more stable.

\begin{figure}[htbp!]   
    \centering
    \includegraphics[width=\linewidth,scale=0.5]{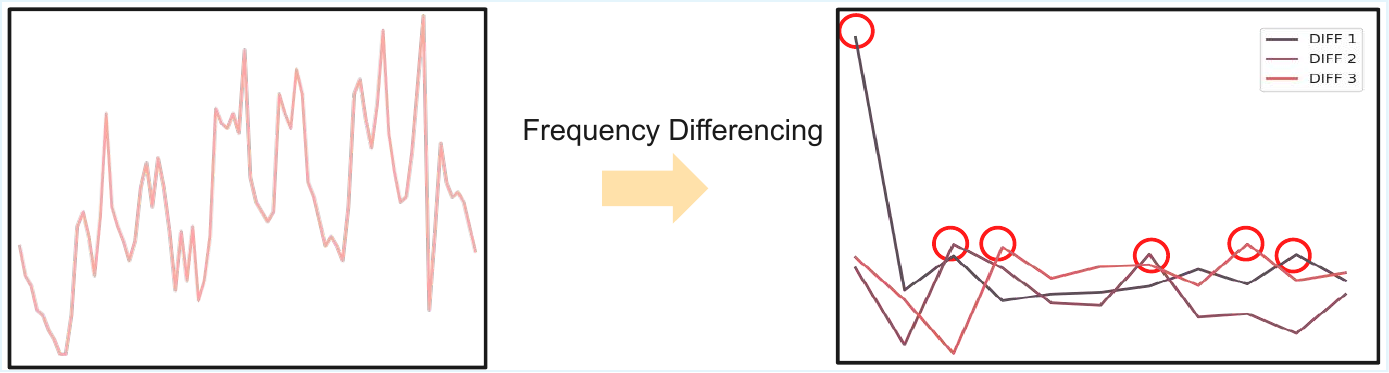}
    \caption{TopK selection Analysis.}
    \label{fig:DIFF_case}
\end{figure}

\textbf{TopK selection Analysis.}
To analyze whether the low-frequency and high-energy components would dominate the selection of TopK, we conduct an experimental analysis. As shown in the left figure of \Cref{fig:DIFF_case}, it is a sampled time series. We divided it into four patches and aligned them for frequency-domain differential operations, obtaining three differential spectrograms as shown in the right figure of Figure 5. We found that for the selection of TopK in the differential spectrograms, it is not dominated by low-frequency and high-energy components. It can be seen that in DIFF 1, low-frequency and high-frequency components are selected, while in DIFF 2 and DIFF 3, the intermediate frequency and high-frequency components are chosen. From this, it can be seen that the model's selection of frequencies is diverse and is not dominated by low-frequency and high-energy components. Due to space limitations, Top-K Frequency, Efficiency, Input Length, and Patch Length Analysis are provided in Appendix B.3-B.6.

\section{Conclusion}
In this paper, we propose a novel model \textbf{DTAF}, which effectively addresses the non-stationarity in both the temporal and frequency domains for time series forecasting. Specifically, our proposed Temporal Stabilizing Fusion (TFS) module is enhanced by a Non-stationary MOE Filter that automatically removes non-stationary patterns in the temporal domain and facilitates the capture of long-term dependencies. Meanwhile, the Frequency Wave Modeling (FWM) module performs frequency domain differencing to handle non-stationarity in the spectral space, allowing the model to better understand periodic trends and sudden shifts. Based on these, DTAF achieves more robust and accurate forecasting across diverse scenarios.

\section{Acknowledgments}

This work was partially supported by the National Natural Science Foundation of China (62372179, 62406112), Scientific and Technological Program of Xi’an under Grant 24GXFW0016.

\bibliography{aaai2026}

\appendix
\section{A Preliminary}

\subsection{A.1 Multivariate Time Series Forecasting}
\label{def:MTSF}

Multivariate Time Series Forecasting (MTSF) refers to the task of predicting future values of multiple channels over a sequence of future timestamps based on their historical observations. Formally, given a multivariate time series \( \mathbf{X} \in \mathbb{R}^{N \times T} \) (with \( N > 1 \) channels and \( T \) historical sequence length), the goal is to learn a function \( \mathrm{f} \) that maps the input \( \mathbf{X} \) to future values \( \mathbf{Y} \in \mathbb{R}^{N \times F} \), where \( F \) is the forecast sequence length.

\subsection{A.2 Differencing}

Differencing is a processing operation usually used in time series analysis to eliminate non-stationarity \cite{bookTSA}. For a multivariate time series \( \mathbf{X} \in \mathbb{R}^{N \times T} \), the first-order differencing is defined as \( \nabla \mathbf{X}_\mathrm{t} = \mathbf{X}_\mathrm{t} - \mathbf{X}_{t-1} \). Higher-order differencing recursively applies this operation (e.g., second-order: \( \nabla^2 \mathbf{X}_\mathrm{t} = \nabla \mathbf{X}_\mathrm{t} - \nabla \mathbf{X}_{t-1} \)), stripping lower-order temporal patterns to stabilize the series and enhance model performance. However, in this paper, we adapt differencing to the frequency domain, enabling the model to capture dynamic patterns in the frequency domain.

\section{B Experiment Details}

\subsection{B.1 Dataset}
\label{Appendix: dataset}

We select 11 widely used time-series datasets to validate the effectiveness of our model, covering a range of domains such as electricity (ETTh1, ETTh2, ETTm1, ETTm2, Electricity), health (ILI, Covid-19), traffic (Traffic), web (Wike2000), environment (Weather), and banking (NN5). Detailed statistics of these datasets are provided in the following.

\begin{table}[htbp!]
\label{Multivariate datasets}
\centering
\resizebox{1\columnwidth}{!}{
\scalebox{1.2}{
\begin{tabular}{@{}lllrrcl@{}}
\toprule
Dataset      & Domain      & Frequency & Timestamps & Channels & Split \\ \midrule
NN5         & Banking     & 1 day     & 791        & 111      & 7:1:2 \\
ILI          & Health      & 1 week     & 966        & 7        & 7:1:2 \\
Covid-19     & Health      & 1 day     & 1,392       & 948      & 7:1:2 \\ 
ETTh1        & Electricity & 1 hour     & 14,400      & 7        & 6:2:2 \\
ETTh2        & Electricity & 1 hour    & 14,400      & 7        & 6:2:2 \\
ETTm1        & Electricity & 15 mins   & 57,600      & 7        & 6:2:2 \\
ETTm2        & Electricity & 15 mins   & 57,600      & 7        & 6:2:2 \\
Electricity  & Electricity & 1 hour    & 26,304      & 321      & 7:1:2 \\
Weather      & Environment & 10 mins   & 52,696      & 21       & 7:1:2 \\
Traffic      & Traffic     & 1 hour    & 17,544      & 862      & 7:1:2 \\
Wike2000    & Web         & 1 day     & 792        & 2,000     & 7:1:2 \\ \bottomrule
\end{tabular}}}
\caption{ap: Statistics of 11 multivariate datasets.}
\end{table}

\subsection{B.2 Baselines}
\label{ap: baselines}
We select several state-of-the-art models as baselines and evaluate them in TFB \cite{tfb}, following TFB's settings \cite{tfb,qiu2025tab}; we do not use the "drop last" trick during the testing phase to ensure a fair comparison. including Amplifier \cite{Amplifier}, which is a model that improves the prediction performance by strengthening the learning of low-energy components; Non-stationary Transformer (Stationary) \cite{stationary}, a variant designed to address non-stationarity by transforming inputs to reduce non-stationarity and stabilize temporal dynamics; iTransformer \cite{itransformer}, a model that enhances traditional Transformer architectures with advanced tokenization and attention mechanisms to capture long-range dependencies; Pathformer \cite{pathformer}, which employs a multi-scale Transformer design with adaptive pathways to effectively integrate multi-resolution information; 
PatchTST \cite{PatchTST}, which segments the time series into patches to reduce complexity while preserving key features; Crossformer \cite{crossformer}, a model that introduces a cross-dimensional attention mechanism to capture inter-channel interactions in multivariate series; FITS \cite{FITS}, a frequency-enhanced model that decomposes the input signal into spectral components to better capture periodic patterns and trends. TimeMixer \cite{timemixer}, leveraging an MLP-Mixer framework to perform temporal mixing without relying on conventional self-attention; TimesNet \cite{timesnet}, which combines time series decomposition with deep neural network architectures to model both local and global dependencies, and DLinear \cite{FC}, a simple yet powerful linear model focusing on capturing linear trends with competitive forecasting performance. The specific code repositories for each of these models are as follows:

\begin{itemize}
    \item Amplifier: \url{https://github.com/aikunyi/Amplifier}
    \item iTransformer: \url{https://github.com/thuml/iTransformer}
    \item Pathformer:\url{https://github.com/decisionintelligence/pathformer}
    \item FITS: \url{https://github.com/VEWOXIC/FITS}
    \item TimeMixer: \url{https://github.com/kwuking/TimeMixer}
    \item PatchTST: \url{https://github.com/yuqinie98/PatchTST}
    \item Crossformer:\url{https://github.com/Thinklab-SJTU/Crossformer}
    \item TimesNet: \url{https://github.com/thuml/TimesNet}
    \item DLinear: \url{https://github.com/cure-lab/LTSF-Linear}
    \item Stationary:\url{https://github.com/thuml/Nonstationary_Transformers}
\end{itemize}

\subsection{B.3 TopK Frequency Analysis.} DTAF adaptively selects the top K frequencies with the greatest variations to adapt to different time series samples. We evaluated the influence of different K values on the prediction accuracy in \Cref{tab: Parameter sensitivity study.}. Our results demonstrate that tracking a broader range of changing frequencies enhances prediction accuracy. In particular, datasets with weaker non-stationary characteristics require capturing more frequency changes to achieve better performance.

\begin{figure*}[htbp!]   
    \centering
    \includegraphics[width=\linewidth,scale=1.0]{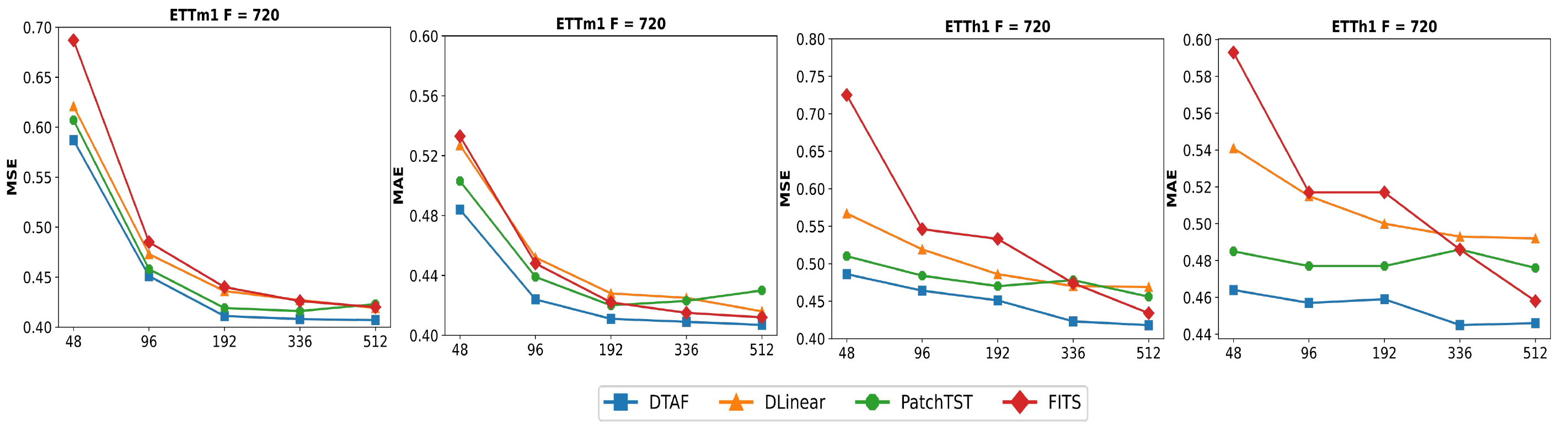}
    \caption{Results with different input length for ETTh1, ETTm1.}
    \label{fig:length}
\end{figure*}

\begin{table}[htbp!]
\centering
\scalebox{0.7}{
\begin{tabular}{cc|cccccccccc}
\toprule
  \multicolumn{2}{c|}{\textbf{TopK}} & \multicolumn{2}{c}{\textbf{K = 2}} & \multicolumn{2}{c}{\textbf{K = 3}} & \multicolumn{2}{c}{\textbf{K = 4}} & \multicolumn{2}{c}{\textbf{K = 5}} \\
\multicolumn{2}{c|}{Metrics}  & mse & mae & mse & mae & mse & mae & mse & mae\\

\midrule
\multirow[c]{4}{*}{\rotatebox{90}{ETTm1}} & 96 & 0.284 & 0.331  & 0.286 & 0.333 & \textbf{0.282} & \textbf{0.330} & 0.285 & 0.331 \\ 
 & 192 & 0.333 & 0.358 & 0.329 & 0.359 & \textbf{0.325} & \textbf{0.358} & 0.328 & 0.358 \\ 
 & 336 & 0.359 & 0.376 & 0.359 & 0.375 & \textbf{0.357} & 0.376 & 0.358 & \textbf{0.375} \\ 
 & 720 & \textbf{0.400} & \textbf{0.410} & 0.411 & 0.410 & 0.415 & 0.410 & 0.418 & 0.410  \\ 
\addlinespace\cline{1-10} \addlinespace

\multirow[c]{4}{*}{\rotatebox{90}{Covid-19}} & 24 & \textbf{0.885} & \textbf{0.0305}  & 0.889 & 0.0310 & 0.890 & 0.0309 & 0.888 & 0.0308 \\ 
 & 36 & 1.186 & 0.0375 & 1.199 & 0.0384 & 1.187 & 0.0380 & \textbf{1.159} & \textbf{0.0372} \\ 
 & 48 & \textbf{1.479} & \textbf{0.0454} & 1.495 & 0.0473 & 1.479 & 0.0457 & 1.498 & 0.0471 \\ 
 & 60 & 1.888 & 0.0513 & \textbf{1.883} & \textbf{0.0510} & 1.891 & 0.0516 & 1.946 & 0.0548  \\ 

\bottomrule
\end{tabular}
}
\caption{Parameter sensitivity study. The MSE and MAE vary with the hyperparameter K, which is used for the TopK selection of frequency.}
\label{tab: Parameter sensitivity study.}
\end{table}

\subsection{B.4 Efficiency Analysis}

\Cref{tab: Efficiency Analysis} compares four time‑series forecasting models on the Weather and ETTh1 datasets across four prediction horizons (96, 192, 336, 720) regarding inference time and parameter count. Pathformer achieves almost the fastest inference in all settings but comes with a relatively large parameter count; DTAF maintains the smallest model size and shows relatively stable inference times; TimesNet, although having the fewest parameters on ETTh1, has more parameters and time costs on Weather, indicating that its parameter count varies with dataset size. Crossformer, while delivering steady inference speeds, has the largest parameter count and thus requires the most computational resources. 

\begin{table}[htbp!]
\centering
\resizebox{1\columnwidth}{!}{
\scalebox{1.2}{
\begin{tabular}{cc|cc|cc|cc|cc|}
\toprule
\multicolumn{2}{c|}{\multirow{2}{*}{Models}} & \multicolumn{2}{c}{DTAF} & \multicolumn{2}{c}{Pathformer} & \multicolumn{2}{c}{TimesNet} & \multicolumn{2}{c}{Crossformer} \\
\multicolumn{2}{c|}{} & \multicolumn{2}{c}{(ours)} & \multicolumn{2}{c}{(2024)} & \multicolumn{2}{c}{(2023)} & \multicolumn{2}{c}{(2023)} \\
\addlinespace\cline{1-10} \addlinespace
\multicolumn{2}{c|}{Metrics} & time & size & time & size & time & size & time & size \\
\midrule

\multirow[c]{4}{*}{\rotatebox{90}{Weather}} & 96 & 3.306 & \textbf{0.420M} & \textbf{2.244} & 5.396M & 3.557 & 37.521M & \underline{2.452} & \underline{2.287M}  \\ 
 & 192 & 2.372 & \textbf{0.814M} & \underline{2.152} & 5.655M & \textbf{1.843} & \underline{1.203M} & 2.433 & 11.364M \\ 
 & 336 & \textbf{1.823} & \textbf{1.414M} & \underline{2.307} & 6.042M & 3.501 & 9.421M & 2.565 & \underline{2.208M} \\ 
 & 720 & \underline{3.194} & \textbf{2.972M} & \textbf{2.900} & \underline{7.074M} & 31.943 & 9.459M & \underline{4.236} & 11.787M \\ 
\addlinespace\cline{1-10} \addlinespace

\multirow[c]{4}{*}{\rotatebox{90}{ETTh1}} & 96 & 1.980 & \textbf{0.421M} & \textbf{1.078} & 0.933M & 1.994 & \underline{0.605M} & \underline{1.603} & 2.686M  \\ 
 & 192 & \underline{1.243} & \underline{0.800M} & \textbf{1.219} & 1.000M & 2.432 & \textbf{0.614M} & 2.059 & 11.336M \\ 
 & 336 & 2.629 & 1.404M & \textbf{1.062} & \underline{1.097M} & \underline{2.598} & \textbf{0.628M} & 2.187 & 11.408M \\ 
 & 720 & \underline{1.852} & 2.972M & \textbf{1.375} & \underline{1.353M} & 2.813 & \textbf{0.666M} & 4.075 & 11.600M \\

\bottomrule
\end{tabular}
}}
\caption{Efficiency Analysis. The metric time represents the reasoning time of the model, and the metric size represents the number of parameters of the model.}
\label{tab: Efficiency Analysis}
\end{table}

\subsection{B.5 Input Length Analysis}

In time series forecasting, the input length decides how much historical information the model uses. We chose some better-performing models from the main experiments as baselines. We compare different input lengths on the ETTh1 and ETTm1 datasets. As shown in \Cref{fig:length}, when the input length increases from 48 to 512, the prediction errors generally decrease. This shows that longer input helps improve forecasting performance. Among all models, our method shows the lowest error across almost all input lengths. These results prove that our model is good at using longer historical data to make better predictions.

\subsection{B.6 Patch Length Analysis}

For the patch length analysis, we conducted hyperparameter sensitivity experiments on it for analysis. Four prediction steps of the ETTh1 dataset were selected for the experiment. We found that for different prediction steps, the selected patch lengths were different; that is, the required local patch features were different. For different prediction lengths, an appropriate patch length needs to be selected to enable the model to correctly extract the required features.

\begin{table}[htbp!]
\centering
\scalebox{0.7}{
\begin{tabular}{cc|cccccccccc}
\toprule
  \multicolumn{2}{c|}{\textbf{Length}} & \multicolumn{2}{c}{\textbf{L = 8}} & \multicolumn{2}{c}{\textbf{L = 16}} & \multicolumn{2}{c}{\textbf{L = 32}} & \multicolumn{2}{c}{\textbf{L = 48}} \\
\multicolumn{2}{c|}{Metrics}  & mse & mae & mse & mae & mse & mae & mse & mae\\

\midrule
\multirow[c]{4}{*}{\rotatebox{90}{ETTh1}} & 96 & \textbf{0.359} & \textbf{0.384}  & 0.363 & 0.386 & 0.361 & 0.385 & 0.365 & 0.386 \\ 
 & 192 & 0.401 & 0.415 & 0.401 & 0.417 & \textbf{0.400} & \textbf{0.414} & 0.407 & 0.418 \\ 
 & 336 & 0.424 & 0.429 & \textbf{0.415} & \textbf{0.424} & 0.419 & 0.424 & 0.419 & 0.430 \\ 
 & 720 & 0.427 & 0.452 & \textbf{0.421} & \textbf{0.449} & 0.428 & 0.453 & 0.439 & 0.460  \\ 

\bottomrule
\end{tabular}
}
\label{tab: Patch Length Analysis}
\caption{Parameter sensitivity study of patch length. The MSE and MAE vary with the hyperparameter patch length.}
\end{table}

\subsection{B.7 Full Main Results}

\Cref{Common Multivariate forecasting results} presents comprehensive multivariate forecasting results, where our model is extensively evaluated against strong baselines across 11 widely-used benchmarks covering both short-term and long-term horizons. These diverse real-world datasets reflect practical forecasting challenges. The results show that our model consistently outperforms state-of-the-art methods in accuracy and robustness to non-stationarity.

\begin{table*}[htbp!]
\resizebox{2\columnwidth}{!}{
\scalebox{1.2}{
\begin{tabular}{cc|cc|cc|cc|cc|cc|cc|cc|cc|cc|cc|cc|}
\toprule
\multicolumn{2}{c|}{\multirow{2}{*}{Models}} & \multicolumn{2}{c}{DTAF} & \multicolumn{2}{c}{Amplifier} & \multicolumn{2}{c}{iTransformer} & \multicolumn{2}{c}{Pathformer} & \multicolumn{2}{c}{FITS} & \multicolumn{2}{c}{TimeMixer} & \multicolumn{2}{c}{PatchTST} & \multicolumn{2}{c}{Crossformer} & \multicolumn{2}{c}{TimesNet} & \multicolumn{2}{c}{DLinear} & \multicolumn{2}{c}{Stationary} \\
\multicolumn{2}{c|}{} & \multicolumn{2}{c}{(ours)} & \multicolumn{2}{c}{(2025)} & \multicolumn{2}{c}{(2024)} & \multicolumn{2}{c}{(2024)} & \multicolumn{2}{c}{(2024)} & \multicolumn{2}{c}{(2024)} & \multicolumn{2}{c}{(2023)} & \multicolumn{2}{c}{(2023)} & \multicolumn{2}{c}{(2023)} & \multicolumn{2}{c}{(2023)} & \multicolumn{2}{c}{(2022)} \\
\addlinespace\cline{1-24} \addlinespace
\multicolumn{2}{c|}{Metrics} & mse & mae & mse & mae & mse & mae & mse & mae & mse & mae & mse & mae & mse & mae & mse & mae & mse & mae & mse & mae & mse & mae \\
\midrule

\multirow[c]{4}{*}{\rotatebox{90}{ILI}} & 24 & \textbf{1.733} & \textbf{0.784} & 1.947 & 0.877 & \underline{1.783} & 0.846 & 2.086 & 0.922 & 2.182 & 1.002 & 1.804 & \underline{0.820} & 1.932 & 0.872 & 2.981 & 1.096 & 2.131 & 0.958 & 2.208 & 1.031 & 2.394 & 1.066 \\ 
 & 36 & \textbf{1.693} & \textbf{0.807} & \underline{1.739} & 0.863 & 1.746 & \underline{0.860} & 1.912 & 0.882 & 2.241 & 1.029 & 1.891 & 0.926 & 1.869 & 0.866 & 3.549 & 1.196 & 2.612 & 0.974 & 2.032 & 0.981 & 2.227 & 1.031 \\ 
 & 48 & \textbf{1.614} & \textbf{0.782} & 1.806 & 0.907 & \underline{1.716} & 0.898 & 1.985 & 0.905 & 2.272 & 1.036 & 1.752 & \underline{0.866} & 1.891 & 0.883 & 3.851 & 1.288 & 1.916 & 0.897 & 2.209 & 1.063 & 2.525 & 1.003 \\ 
 & 60 & \textbf{1.714} & \textbf{0.832} & \underline{1.786} & 0.907 & 2.183 & 0.963 & 1.999 & 0.929 & 2.642 & 1.142 & 1.831 & 0.930 & 1.914 & \underline{0.896} & 4.692 & 1.450 & 1.995 & 0.905 & 2.292 & 1.086 & 2.410 & 1.010 \\ 
\addlinespace\cline{1-24} \addlinespace

\multirow[c]{4}{*}{\rotatebox{90}{Covid-19}} & 24 & \textbf{0.885} & \textbf{0.030} & 3.691 & 0.088 & \underline{1.001} & \underline{0.038} & 1.424 & 0.058 & 4.934 & 0.104 & 2.781 & 0.087 & 1.045 & 0.042 & 1768.817 & 2.314 & 2.215 & 0.064 & 9.587 & 0.240 & 2.234 & 0.062 \\ 
 & 36 & \textbf{1.159} & \textbf{0.037} & 4.149 & 0.100 & \underline{1.236} & \underline{0.042} & 3.109 & 0.085 & 4.935 & 0.113 & 2.877 & 0.093 & 1.397 & 0.051 & 1770.939 & 2.346 & 2.531 & 0.081 & 5.333 & 0.173 & 2.408 & 0.076 \\ 
 & 48 & \textbf{1.479} & \textbf{0.045} & 6.155 & 0.118 & \underline{1.710} & \underline{0.056} & 4.543 & 0.096 & 5.927 & 0.131 & 9.916 & 0.138 & 1.769 & 0.062 & 1773.447 & 2.450 & 2.672 & 0.079 & 9.598 & 0.246 & 2.705 & 0.083 \\ 
 & 60 & \textbf{1.883} & \textbf{0.051} & 8.317 & 0.144 & \underline{2.005} & \underline{0.062} & 4.227 & 0.105 & 6.880 & 0.145 & 36.191 & 0.215 & 2.216 & 0.068 & 1772.833 & 2.486 & 3.018 & 0.085 & 7.778 & 0.297 & 3.285 & 0.093 \\ 
\addlinespace\cline{1-24} \addlinespace

\multirow[c]{4}{*}{\rotatebox{90}{NN5}} & 24 & \textbf{0.716} & \textbf{0.559} & 0.918 & 0.685 & 0.727 & \underline{0.568} & 0.769 & 0.602 & 0.870 & 0.663 & \underline{0.723} & 0.723 & 0.740 & 0.596 & 0.741 & 0.591 & 0.739 & 0.579 & 0.752 & 0.594 & 1.274 & 0.900 \\ 
 & 36 & \textbf{0.644} & \textbf{0.537} & 1.392 & 0.897 & 0.664 & \underline{0.552} & 0.701 & 0.583 & 0.814 & 0.655 & \underline{0.657} & 0.557 & 0.694 & 0.595 & 0.703 & 0.589 & 0.717 & 0.585 & 0.690 & 0.578 & 1.318 & 0.930 \\ 
 & 48 & \textbf{0.621} & \textbf{0.535} & 2.196 & 1.170 & 0.633 & \underline{0.543} & 0.668 & 0.573 & 0.780 & 0.644 & \underline{0.630} & 0.550 & 0.667 & 0.585 & 0.669 & 0.575 & 0.647 & 0.558 & 0.665 & 0.573 & 1.277 & 0.905 \\ 
 & 60 & \textbf{0.593} & \textbf{0.522} & 2.671 & 1.322 & 0.615 & \underline{0.537} & 0.655 & 0.570 & 0.781 & 0.650 & \underline{0.612} & 0.543 & 0.653 & 0.582 & 0.683 & 0.587 & 0.633 & 0.547 & 0.659 & 0.575	& 1.313 & 0.927 \\ 
\addlinespace\cline{1-24} \addlinespace

\multirow[c]{4}{*}{\rotatebox{90}{Wike2000}} & 24 & \textbf{428.048} & \textbf{0.982} & 565.072 & 1.210 & 453.475 & \underline{1.011} & 664.769 & 1.357 & 669.429 & 1.351 & \underline{447.039} & 1.079 & 457.185 & 1.023 & 638.794 & 1.734 & 521.367 & 1.215 & 556.014 & 1.274 & 845.927 & 1.423 \\ 
 & 36 & \textbf{483.089} & \textbf{1.094} & 612.685 & 1.294 & 515.830 & 1.132 & 699.269 & 1.431 & 711.479 & 1.440 & 510.073 & 1.221 & 511.943 & \underline{1.115} & 692.485 & 1.808 & \underline{506.339} & 1.201 & 613.963 & 1.355 & 1326.986 & 1.595 \\ 
 & 48 & \textbf{509.093} & \textbf{1.161} & 631.723 & 1.362 & 578.335 & 1.214 & 847.933 & 1.604 & 757.815 & 1.537 & 542.797 & 1.303 & 531.901 & \underline{1.179} & 718.072 & 1.899 & \underline{523.611} & 1.251 & 659.957 & 1.437 & 720.409 & 1.406 \\ 
 & 60 & \textbf{545.186} & \textbf{1.257} & 857.566 & 1.602 & 634.947	 & 1.402 & 942.510 & 1.717 & 791.442 & 1.615 & 574.808 & 1.402 & \underline{554.829} & \underline{1.244} & 730.368 & 2.021 & 559.841 & 1.329	 & 693.140 & 1.510 & 1060.171 & 1.578 \\ 
\addlinespace\cline{1-24} \addlinespace

\multirow[c]{4}{*}{\rotatebox{90}{ETTh1}} & 96 & \textbf{0.359} & \textbf{0.384} & \underline{0.371} & \underline{0.392} & 0.386 & 0.405 & 0.372 & 0.392 & 0.376 & 0.396 & 0.372 & 0.401 & 0.377 & 0.397 & 0.411 & 0.435 & 0.389 & 0.412 & 0.379 & 0.403 & 0.591 & 0.524 \\ 
 & 192 & \textbf{0.400} & \textbf{0.414} & 0.409 & 0.424 & 0.424 & 0.440 & 0.408 & \underline{0.415} & \underline{0.400} & 0.418 & 0.413 & 0.430 & 0.409 & 0.425 & 0.409 & 0.438 & 0.440 & 0.443 & 0.408 & 0.419 & 0.615 & 0.540 \\ 
 & 336 & \textbf{0.415} & \textbf{0.424} & 0.443 & 0.448 & 0.449 & 0.460 & 0.438 & \underline{0.434} & \underline{0.419} & 0.435 & 0.438 & 0.450 & 0.431 & 0.444 & 0.433 & 0.457 & 0.523 & 0.487 & 0.440 & 0.440 & 0.632 & 0.551 \\ 
 & 720 & \textbf{0.421} & \textbf{0.449} & 0.443 & 0.461 & 0.495 & 0.487 & 0.450 & 0.463 & \underline{0.435} & \underline{0.458} & 0.486 & 0.484 & 0.457 & 0.477 & 0.501 & 0.514 & 0.521 & 0.495 & 0.471 & 0.493 & 0.828 & 0.658 \\ 
\addlinespace\cline{1-24} \addlinespace

\multirow[c]{4}{*}{\rotatebox{90}{ETTh2}} & 96 & \textbf{0.271} & \textbf{0.329} & 0.292 & 0.354 & 0.297 & 0.348 & 0.279 & \underline{0.336} & 0.277 & 0.345 & 0.281 & 0.351 & \underline{0.274} & 0.337 & 0.728 & 0.603 & 0.334 & 0.370 & 0.300 & 0.364 & 0.347 & 0.387 \\ 
 & 192 & \textbf{0.330} & \textbf{0.367} & 0.346 & 0.390 & 0.372 & 0.403 & 0.345 & 0.380 & \underline{0.331} & \underline{0.379} & 0.349 & 0.387 & 0.348 & 0.384 & 0.723 & 0.607 & 0.404 & 0.413 & 0.387 & 0.423 & 0.379 & 0.418 \\ 
 & 336 & \underline{0.364} & \textbf{0.395} & 0.378 & 0.425 & 0.388 & 0.417 & 0.378 & 0.408 & \textbf{0.350} & \underline{0.396} & 0.366 & 0.413 & 0.377 & 0.416 & 0.740 & 0.628 & 0.389 & 0.435 & 0.490 & 0.487 & 0.358 & 0.413 \\ 
 & 720 & \underline{0.392} & \textbf{0.424} & 0.402 & 0.440 & 0.424 & 0.444 & 0.437 & 0.455 & \textbf{0.382} & \underline{0.425} & 0.401 & 0.436 & 0.406 & 0.441 & 1.386 & 0.882 & 0.434 & 0.448 & 0.704 & 0.597 & 0.422 & 0.457 \\ 
\addlinespace\cline{1-24} \addlinespace

\multirow[c]{4}{*}{\rotatebox{90}{ETTm1}} & 96 & \textbf{0.282} & \textbf{0.330} & 0.313 & 0.359 & 0.300 & 0.353 & 0.290 & \underline{0.335} & 0.303 & 0.345 & 0.293 & 0.345 & \underline{0.289} & 0.343 & 0.314 & 0.367 & 0.340 & 0.378 & 0.300 & 0.345 & 0.415 & 0.410 \\ 
 & 192 & \textbf{0.325} & \textbf{0.358} & 0.350 & 0.378 & 0.341 & 0.380 & 0.337 & \underline{0.363} & 0.337 & 0.365 & 0.335 & 0.372 & \underline{0.329} & 0.368 & 0.374 & 0.410 & 0.392 & 0.404 & 0.336 & 0.366 & 0.494 & 0.451 \\ 
 & 336 & \textbf{0.357} & \textbf{0.376} & 0.376 & 0.392 & 0.374 & 0.396 & 0.374 & 0.384 & 0.368 & \underline{0.384} & 0.368 & 0.386 & \underline{0.362} & 0.390 & 0.413 & 0.432 & 0.423 & 0.426 & 0.367 & 0.386 & 0.577 & 0.490 \\ 
 & 720 & \textbf{0.407} & \textbf{0.407} & 0.459 & 0.439 & 0.429 & 0.430 & 0.428 & 0.416 & 0.420 & \underline{0.413} & 0.426 & 0.417 & \underline{0.416} & 0.423 & 0.753 & 0.613 & 0.475 & 0.453 & 0.419 & 0.416 & 0.636 & 0.535 \\ 
\addlinespace\cline{1-24} \addlinespace

\multirow[c]{4}{*}{\rotatebox{90}{ETTm2}} & 96 & \textbf{0.161} & \textbf{0.246} & 0.184 & 0.271 & 0.175 & 0.266 & \underline{0.164} & \underline{0.250} & 0.165 & 0.254 & 0.165 & 0.256 & 0.165 & 0.255 & 0.296 & 0.391 & 0.189 & 0.265 & 0.164 & 0.255 & 0.210 & 0.294 \\ 
 & 192 & \textbf{0.215} & \textbf{0.284} & 0.248 & 0.312 & 0.242 & 0.312 & \underline{0.219} & \underline{0.288} & 0.219 & 0.291 & 0.225 & 0.298 & 0.221 & 0.293 & 0.369 & 0.416 & 0.254 & 0.310 & 0.224 & 0.304 & 0.338 & 0.373 \\ 
 & 336 & 0.269 & \textbf{0.318} & 0.321 & 0.362 & 0.282 & 0.337 & \textbf{0.267} & \underline{0.319} & \underline{0.272} & 0.326 & 0.277 & 0.332 & 0.276 & 0.327 & 0.588 & 0.600 & 0.313 & 0.345 & 0.277 & 0.337 & 0.432 & 0.416 \\ 
 & 720 & 0.367 & \underline{0.384} & 0.374 & 0.400 & 0.375 & 0.394 & 0.361 & \textbf{0.377} & \textbf{0.359} & 0.381 & \underline{0.360} & 0.387 & 0.362 & 0.381 & 0.750 & 0.612 & 0.413 & 0.402 & 0.371 & 0.401 & 0.554 & 0.476 \\ 
\addlinespace\cline{1-24} \addlinespace

\multirow[c]{4}{*}{\rotatebox{90}{Weather}} & 96 & \underline{0.145} & \textbf{0.185} & 0.146 & 0.201 & 0.157 & 0.207 & 0.148 & \underline{0.195} & 0.172 & 0.225 & 0.147 & 0.198 & 0.149 & 0.196 & \textbf{0.143} & 0.210 & 0.168 & 0.214 & 0.170 & 0.230 & 0.188 & 0.242 \\ 
 & 192 & \textbf{0.190} & \textbf{0.228} & \underline{0.190} & 0.240 & 0.200 & 0.248 & 0.191 & \underline{0.235} & 0.215 & 0.261 & 0.192 & 0.243 & 0.191 & 0.239 & 0.198 & 0.260 & 0.219 & 0.262 & 0.216 & 0.273 & 0.241 & 0.290 \\ 
 & 336 & \textbf{0.241} & \textbf{0.269} & \underline{0.241} & 0.281 & 0.252 & 0.287 & 0.243 & \underline{0.274} & 0.261 & 0.295 & 0.247 & 0.284 & 0.242 & 0.279 & 0.258 & 0.314 & 0.278 & 0.302 & 0.258 & 0.307 & 0.341 & 0.341 \\ 
 & 720 & \underline{0.313} & \textbf{0.321} & 0.313 & 0.330 & 0.320 & 0.336 & 0.318 & \underline{0.326} & 0.326 & 0.341 & 0.318 & 0.330 & \textbf{0.312} & 0.330 & 0.335 & 0.385 & 0.353 & 0.351 & 0.323 & 0.362 & 0.403 & 0.388 \\ 
\addlinespace\cline{1-24} \addlinespace

\multirow[c]{4}{*}{\rotatebox{90}{Electricity}} & 96 & \textbf{0.129} & \textbf{0.221} & 0.141 & 0.238 & \underline{0.134} & 0.230 & 0.135 & \underline{0.222} & 0.139 & 0.237 & 0.153 & 0.256 & 0.143 & 0.247 & 0.134 & 0.231 & 0.169 & 0.271 & 0.140 & 0.237 & 0.171 & 0.274 \\ 
 & 192 & \textbf{0.146} & \textbf{0.235} & 0.164 & 0.257 & \underline{0.154} & \underline{0.250} & 0.157 & 0.253 & 0.154 & 0.250 & 0.168 & 0.269 & 0.158 & 0.260 & 0.146 & 0.243 & 0.180 & 0.280 & 0.154 & 0.251 & 0.180 & 0.283 \\ 
 & 336 & \textbf{0.165} & \textbf{0.254} & 0.188 & 0.280 & 0.169 & \underline{0.265} & 0.170 & 0.267 & 0.170 & 0.268 & 0.189 & 0.291 & \underline{0.168} & 0.267 & 0.165 & 0.264 & 0.204 & 0.304 & 0.169 & 0.268 & 0.204 & 0.305 \\ 
 & 720 & \underline{0.200} & \textbf{0.283} & 0.204 & 0.293 & \textbf{0.194} & \underline{0.288} & 0.211 & 0.302 & 0.212 & 0.304 & 0.228 & 0.320 & 0.214 & 0.307 & 0.237 & 0.314 & 0.205 & 0.304 & 0.204 & 0.301 & 0.221 & 0.319 \\ 
\addlinespace\cline{1-24} \addlinespace

\multirow[c]{4}{*}{\rotatebox{90}{Traffic}} & 96 & 0.370 & \textbf{0.232} & 0.399 & 0.283 & \textbf{0.363} & 0.265 & 0.384 & \underline{0.250} & 0.400 & 0.280 & \underline{0.369} & 0.257 & 0.370 & 0.262 & 0.526 & 0.288 & 0.595 & 0.312 & 0.395 & 0.275 & 0.604 & 0.330 \\ 
 & 192 & 0.392 & \textbf{0.243} & 0.413 & 0.287 & \textbf{0.384} & 0.273 & 0.405 & \underline{0.257} & 0.412 & 0.288 & 0.400 & 0.272 & \underline{0.386} & 0.269 & 0.503 & 0.263 & 0.613 & 0.322 & 0.407 & 0.280 & 0.610 & 0.338 \\ 
 & 336 & 0.408 & \textbf{0.252} & 0.424 & 0.295 & \textbf{0.396} & 0.277 & 0.424 & \underline{0.265} & 0.426 & 0.301 & 0.407 & 0.272 & \underline{0.396} & 0.275 & 0.505 & 0.276 & 0.626 & 0.332 & 0.417 & 0.286 & 0.626 & 0.341 \\ 
 & 720 & \underline{0.441} & \textbf{0.269} & 0.458 & 0.311 & 0.445 & 0.308 & 0.452 & \underline{0.283} & 0.478 & 0.339 & 0.461 & 0.316 & \textbf{0.435} & 0.295 & 0.552 & 0.301 & 0.635 & 0.340 & 0.454 & 0.308 & 0.643 & 0.347 \\ 
\addlinespace\cline{1-24} \addlinespace

\multicolumn{2}{c|}{$1^{st}$ Count}& \textbf{33} & \textbf{43} & 0 & 0 & \underline{4} & \underline{0} & 1 & 1 & 3 & 0 & 0 & 0 & 1 & 0 & 2 & 0 & 0 & 0 & 0 & 0 & 0 & 0 \\ 
\addlinespace

\bottomrule
\end{tabular}
}}
\caption{Multivariate forecasting results. The best results are highlighted in bold, and the second-best results are underlined.}
\label{Common Multivariate forecasting results}
\end{table*}

\label{appendix:FR}

\end{document}